\documentclass[conference]{IEEEtran}
\IEEEoverridecommandlockouts
\usepackage{cite}
\usepackage{amsmath,amssymb,amsfonts}
\usepackage{algorithmic}
\usepackage{graphicx}
\usepackage{array}
\usepackage{tabularx}
\usepackage{textcomp}
\usepackage{tabularx}
\usepackage{caption}
\usepackage{subcaption}

\usepackage{xcolor}
\def\BibTeX{{\rm B\kern-.05em{\sc i\kern-.025em b}\kern-.08em
    T\kern-.1667em\lower.7ex\hbox{E}\kern-.125emX}}
\begin{document}

\title{Detecting Facial Image Manipulations with Multi-Layer CNN Models}

\author{
\IEEEauthorblockN{Alejandro Marco Montejano}
\IEEEauthorblockA{
    \textit{Facephi Research Laboratory} \\
    Alicante, Spain \\
    alejandromarco@facephi.com}
\and
\IEEEauthorblockN{Angela Sanchez Perez}
\IEEEauthorblockA{
    \textit{Facephi Research Laboratory} \\
    Alicante, Spain \\
    asanchezperez@facephi.com}
\and
\IEEEauthorblockN{Javier Barrachina}
\IEEEauthorblockA{
    \textit{Facephi Research Laboratory} \\
    Alicante, Spain \\
    jbarrachina@facephi.com}
\and
\IEEEauthorblockN{David Ortiz-Perez}
\IEEEauthorblockA{
    \textit{Dept. of Computer Science and} \\
    \textit{Technology, University of Alicante} \\
    Alicante, Spain \\
    dortiz@dtic.ua.es}
\and
\IEEEauthorblockN{Manuel Benavent-Lledo}
\IEEEauthorblockA{
    \textit{Dept. of Computer Science and} \\
    \textit{Technology, University of Alicante} \\
    Alicante, Spain \\
    mbenavent@dtic.ua.es}
\and
\IEEEauthorblockN{Jose Garcia-Rodriguez}
\IEEEauthorblockA{
    \textit{Dept. of Computer Science and} \\
    \textit{Technology, University of Alicante} \\
    Alicante, Spain \\
    jgarcia@dtic.ua.es}

}

\maketitle

\begin{abstract}
The rapid evolution of digital image manipulation techniques poses significant challenges for content verification, with models such as stable diffusion and mid-journey producing highly realistic, yet synthetic, images that can deceive human perception. This research develops and evaluates convolutional neural networks (CNNs) specifically tailored for the detection of these manipulated images. The study implements a comparative analysis of three progressively complex CNN architectures, assessing their ability to classify and localize manipulations across various facial image modifications. Regularization and optimization techniques were systematically incorporated to improve feature extraction and performance. The results indicate that the proposed models achieve an accuracy of up to 76\% in distinguishing manipulated images from genuine ones, surpassing traditional approaches. This research not only highlights the potential of CNNs in enhancing the robustness of digital media verification tools, but also provides insights into effective architectural adaptations and training strategies for low-computation environments. Future work will build on these findings by extending the architectures to handle more diverse manipulation techniques and integrating multi-modal data for improved detection capabilities.
\end{abstract}


\section{Introduction}

In an increasingly complex digital environment, the authenticity and integrity of visual content are critical for fostering trust and security. The rapid advancement of image generation technologies has made it easier than ever for individuals to create highly realistic visual content, including human representations, objects, and scenes, using simple commands or prompts\cite{dang2020detection}. While this accessibility has unleashed unprecedented creativity, it also raises significant concerns about the veracity of visual media, particularly in fields like journalism, communication, biometric identification, and online security.

Among the most pressing challenges are facial manipulations, which can distort reality and create deceptive imagery. Techniques such as face morphing\cite{medvedev2024}, splicing\cite{xiang2024}, face swapping\cite{Rossler_2019_ICCV}, and subtle alterations to facial expressions can mislead viewers and pose serious consequences. For example, face morphing blends features from multiple individuals into hybrid images, while splicing combines facial elements from distinct images to fabricate false representations\cite{medvedev2024}. Face swapping replaces one face with another, and subtle adjustments to facial gestures or expressions can manipulate perceptions without detection\cite{Rossler_2019_ICCV}. These techniques complicate the verification of image authenticity, underscoring the urgent need for effective detection methods.

These facial manipulation techniques pose substantial challenges for image authentication and verification, especially in contexts where digital manipulation can lead to serious consequences, such as biometric identification and online security\cite{Singh2021}. Therefore, it is essential to develop effective methods for detecting and mitigating these manipulations to ensure the integrity and authenticity of visual content in critical environments, such as news and communication.

To address these challenges, this study focuses on developing tools and methodologies to detect and classify facial manipulations, as well as other forms of digital image tampering. The proposed solutions leverage deep learning models, specifically convolutional neural networks (CNNs), to detect subtle statistical differences between authentic and manipulated images. The study builds upon the MesoNet\cite{mesonet} binary classification model, addressing its key limitations: the inability to generalize beyond its training dataset and its restriction to a single class of image manipulation attacks. By overcoming these constraints, the research aims to enhance the reliability of visual content authentication, particularly in resource-limited environments where computational efficiency is essential.

In summary, the contributions of this paper are the following:

\begin{itemize}
    \item Development of a scalable and efficient multi-class classification system that achieves notable accuracy while maintaining computational efficiency. The system is designed to adapt to diverse requirements and effectively distinguishes between various attack types and genuine images.
    \item Inference was conducted using the MesoNet binary classification model, however, the model demonstrated limited performance in accurately classifying previously unseen images outside the training dataset. To address this limitation, enhancements to the model's architecture were implemented, aiming to improve its generalization and classification capabilities.
    \item Development of a computationally demanding multi-classifier system, MultiMesoNet, to classify between genuine images and different image manipulation attacks, based on a 4 layer convolutional architecture inspired by MesoNet.
    \item Proposal and development of a MultiMesoNet+ 6 layer convolutional multi-classification architecture, with genuine image classification results and different image manipulation attacks with a high level of security.

\end{itemize}

The remainder of the paper is organized as follows. Section~\ref{sec:relwork} summarizes related work in detecting facial image manipulations. Section~\ref{sec:method} describes the methods proposed to address the aforementioned challenges. The experimental setup is detailed in Section~\ref{sec:exp} along with the results. Finally, conclusions from this work are drawn in Section~\ref{sec:conclusion}.

\section{Related Work}\label{sec:relwork}

Previous work in relevant topics is reviewed.

\subsection{Current status of synthetic image generation and detection}

Probabilistic noise diffusion models, particularly Denoising Diffusion Probabilistic Models (DDPM), have emerged as a leading technique in image processing and computer vision (CV) for generating high-quality images from Gaussian noise\cite{ho2020}. This iterative approach gradually incorporates noise during training and subsequently removes it, enhancing the accuracy of generated data.

Generative Adversarial Networks (GANs) have been widely utilized for synthetic image generation, consisting of a generator and a discriminator that compete against each other. The generator learns to create realistic data that are indistinguishable from real samples, while the discriminator aims to identify the authenticity of the data. Although GANs excel at generating diverse and photorealistic images, they face challenges with regard to training stability and control over the generated results.\cite{hong2019}

In contrast, DDPMs offer advantages in training stability, result fidelity, and control during the generation process. They utilize a parameterized Markov chain structure trained via variational inference techniques to produce samples that fit observed data\cite{kim2022}. Recent advances have demonstrated that diffusion models can effectively generate high-quality samples by implementing denoising strategies across multiple noise levels during training, coupled with Langevin dynamics during sampling.

The ID Conditional DDPM has emerged as a key player in diffusion-based face exchange, allowing for the controlled transfer of facial identities while preserving attributes of the target face, such as expression and pose. This model leverages pre-trained expert facial models to ensure high fidelity and integrity in face synthesis.\cite{kim2022}

In addition, transformers, known for their success in natural language processing (NLP), have shown promise in digitally manipulated image detection. By capturing complex relationships in multimodal data, transformer-based models can analyze both images and text to identify digital manipulations, thus aiding in the preservation of visual content authenticity in an increasingly complex digital landscape.\cite{shao2023}

\subsection{Potential of Multimodal Deep Learning Models for Detection and Localization of Manipulated Facial Images}

In the field of machine learning, there are two primary approaches for processing and understanding data: unimodal models, which handle a single source of data, and multimodal models, which integrate multiple sources or modalities, such as audio, video, text, and images. Multimodal models excel in capturing information from various sources and learning complex interrelationships, enabling them to perform specialized tasks effectively\cite{shao2023}.

These models are particularly beneficial in applications where information from multiple sources is interrelated, such as in the detection of digital manipulations. Utilizing multimodal models can significantly enhance the results and performance of these tasks. Within the realm of multimodal learning, two specific learning methods stand out for analyzing and processing data, particularly in the context of text and image interactions\cite{shao2023}.

Human faces are essential for communication and the association of identities, including gender and age. Facial recognition technology is susceptible to manipulation by malicious actors.

To address this issue, an approach employing an attentional mechanism for the detection of manipulated facial images and the identification of specific tampered regions is proposed. By refining feature maps during the classification task, the learned attention maps emphasize informative regions, thereby enhancing the accuracy of distinguishing between authentic and manipulated faces and visualizing tampered areas.

\subsection{MesoNet. A compact network for facial video
forgery detection}

The use of compact networks enables the implementation of detection models on devices with limited computational capacity, facilitating operation without complex systems. DeepFakes can be generated using autoencoders, which compress image data in an encoder to reduce noise and computational cost. The original image can be restored through a decoder. The training process involves extracting faces from existing deepfake videos by selecting specific frames, ensuring that various face angles and resolution levels are uniformly distributed across genuine and deepfake datasets\cite{zhang2024}.

MesoNet, a specialized neural network designed to detect manipulations in facial videos, primarily targets techniques such as Deepfake and Face2Face. It consists of two main architectures: Meso-4 and MesoInception-4. The Meso-4 architecture features four convolutional and pooling layers for effective feature extraction from facial images. The MesoInception-4 architecture enhances this by replacing the first two convolutional layers with a modified inception module that utilizes 3x3 dilated convolutions and 1x1 convolutions to reduce dimensions and improve network performance\cite{zhang2024}.

\subsection{Impact of Convolutional Filter Size and Number on Feature Extraction in the RGB Color Space Using MesoNet}

The Meso4 model is a convolutional neural network architecture comprising four convolutional blocks followed by a fully connected hidden layer. It processes input images of 256x256 pixels in RGB color space, utilizing three color channels. The pre-trained model employs predefined weights to ensure accurate inference on new images\cite{zhang2024}.

Each convolutional block consists of a convolutional layer, a max pooling layer, and a batch normalization layer. The convolutional layer is fundamental, with configurable parameters such as filter size and the number of filters applied. Each filter represents a distinct visual feature (e.g., horizontal or vertical lines) and slides over the input image to assess similarity through matrix operations, calculating the dot product between the filter and specific image regions for each color channel. This enables the neural network to detect relevant patterns and characteristics\cite{zhang2024}.

Based on the results of MesoMultiNet, the goal is to develop a model capable of generalizing and accurately detecting different types of manipulated images. In this new architecture, the number of convolutional layers is increased from 4 to 6, with an expansion in the number of filters and an optimized distribution of layer sizes to improve the detection of edges, fine details, and general characteristics across different classes\cite{zhang2024}.

The final stage of the convolutional blocks features the max pooling layer, which reduces data dimensionality, significantly accelerating computation. Although this may seem like data is discarded, the convolution process allows the model to focus on key features, akin to how the human visual system prioritizes relevant information. Subsequent blocks in the CNN architecture progress toward more complex feature representations, evolving from simple lines to corners, shapes, and faces\cite{zhang2024}.

The size and number of convolutional filters in the MesoNet architecture depend on several key factors. The original MesoNet configuration utilizes four convolutional layers: the first two layers have 8 filters with sizes of 3x3 and 5x5, while the subsequent two layers use 16 filters, both with a size of 3x3. The selection of filter sizes is influenced by the input image dimensions, the complexity of features to capture, and the available computational resources. Commonly used filter sizes, such as 3x3 and 5x5, capture different types of image features\cite{keshari2018}. Using varied filter sizes across layers enables the network to learn a broad spectrum of features at multiple spatial scales, improving its capacity to model complex and discriminative patterns\cite{agrawal2020}.

The 3x3 filters are optimized for capturing fine details and edges, while the 5x5 filters are designed for detecting larger and more complex image patterns\cite{keshari2018}. The number of filters chosen for each convolutional layer affects the model's ability to learn and represent features at varying levels of abstraction. Typical configurations include layers with 8, 16, 32, or 64 filters\cite{agrawal2020}.

Adjusting the number of filters impacts the model's complexity, parameter count, and computational cost. Increasing the number of filters enhances the model's capacity to extract high-level features, which is advantageous for complex datasets. However, this also raises the risk of overfitting if the training data is insufficient. Therefore, careful consideration of the number of filters is necessary to balance feature extraction capabilities and the model's generalization performance.

\section{Methodology}\label{sec:method}

This section presents the methodology employed in this research to address the challenges of detecting and classifying digitally manipulated facial images. Existing lightweight models, such as MesoNet, are advantageous due to their low computational requirements, making them well-suited for resource-constrained environments. However, their binary classification framework and reduced accuracy on unseen datasets reveal notable limitations. To address these issues, this study introduces advanced preprocessing techniques and architectural improvements designed to enhance the model's robustness and generalization capabilities while maintaining computational efficiency. The result is a novel architecture designated as MesoNet+.

\begin{figure}[htpb]
    \centering
    \begin{subfigure}[b]{0.24\textwidth}
        \centering
        \includegraphics[width=3cm, height=3cm, keepaspectratio]{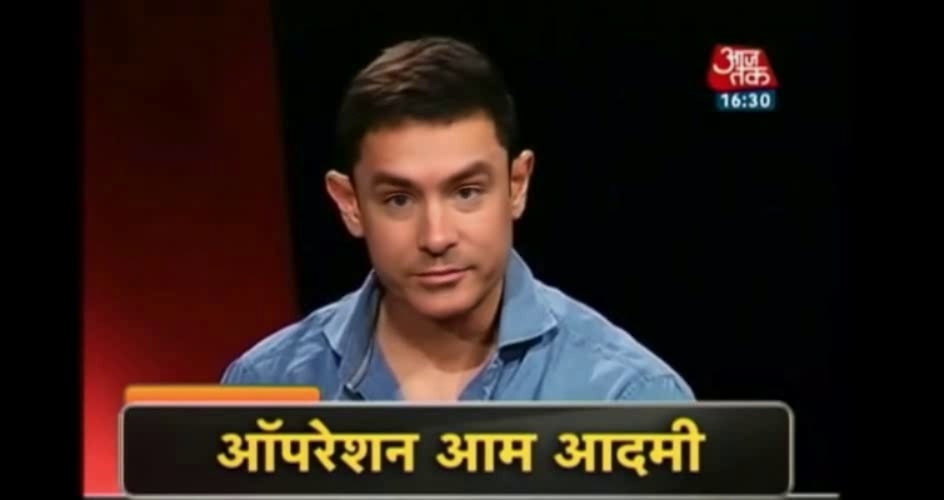}
        \caption{DeepFake - Original}
    \end{subfigure}
    \hfill
    \begin{subfigure}[b]{0.24\textwidth}
        \centering
        \includegraphics[width=3cm, height=3cm, keepaspectratio]{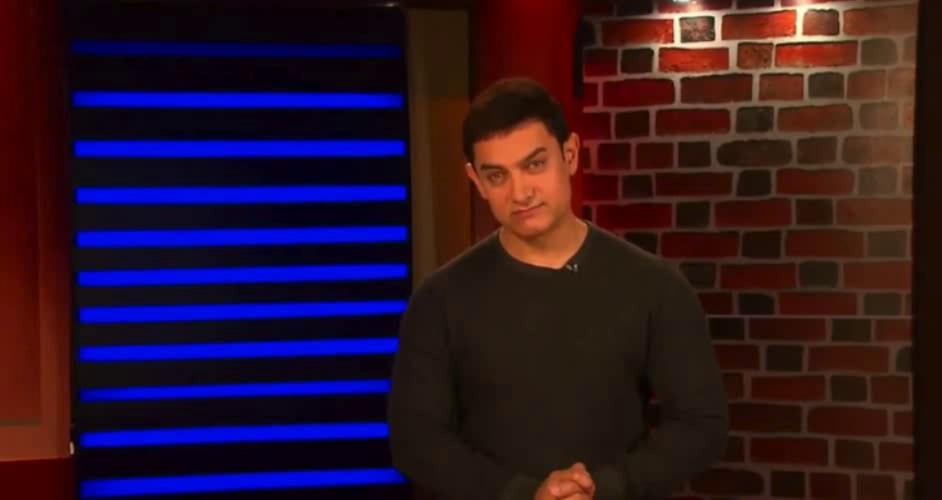}
        \caption{Bonafide - Original}
    \end{subfigure}

    \vspace{0cm} 
    \begin{subfigure}[b]{0.24\textwidth}
        \centering
        \includegraphics[width=3cm, height=3cm, keepaspectratio]{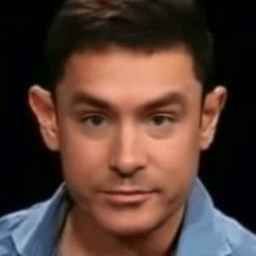}
        \caption{DeepFake - Aligned}
    \end{subfigure}
    \hfill
    \begin{subfigure}[b]{0.24\textwidth}
        \centering
        \includegraphics[width=3cm, height=3cm, keepaspectratio]{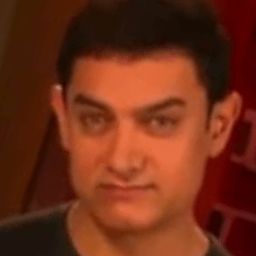}
        \caption{Bonafide - Aligned}
    \end{subfigure}
    
    \caption{Comparison of original and aligned images for DeepFake and Bonafide classes.}
    \label{fig:comparativa_imagenes}
\end{figure}

A critical component in this methodology is the preprocessing and alignment of input images. Proper image alignment ensures standardized face positions and sizes, facilitating feature analysis and enhancing model performance. Facial landmarks, such as eyes, nose, and mouth, are detected using a pre-trained model \cite{dlib_face_landmarks}, enabling geometric transformations like translation, rotation, scaling, and cropping. These transformations ensure consistent orientation and size, optimizing the model's ability to detect anomalies and classify manipulation types accurately. An example of these transformation can be seen in Figure \ref{fig:comparativa_imagenes}.

The first proposed architecture, MesoNet+, is a refined binary classification model based on the original MesoNet. This enhanced design incorporates two additional convolutional layers, increasing the model's capacity to extract and analyze subtle features. These modifications address MesoNet's limitations, particularly its reduced generalization performance on unseen datasets, by refining the attention mechanism and enhancing feature extraction capabilities. 

Building on this foundation, the focus shifts to multi-class classification with the development of the MesoMultiNet model. This architecture adapts the MesoNet framework to classify multiple manipulation types, employing a Softmax activation function for multi-class output. To further improve classification accuracy, the MesoMultiNet+ architecture is introduced. This advanced model features six convolutional layers and additional filters, enabling it to capture finer details and more complex patterns. These iterative improvements, coupled with advanced preprocessing techniques, transfer learning from pre-trained binary models, and carefully designed training processes, yield robust models capable of generalizing effectively across diverse datasets while maintaining computational efficiency.

\subsection{Binary Classification System: MesoNet+}

While MesoNet demonstrates reasonable performance on known datasets, its generalization to unseen samples remains limited. To overcome this limitation, we propose MesoNet+, an enhanced version of the original architecture that incorporates two additional convolutional layers. These layers improve the model's capacity to capture subtle features in manipulated images by increasing the depth and expressiveness of its convolutional filters.

\begin{figure}[htpb]
    \centering
    \includegraphics[width=0.45\textwidth]{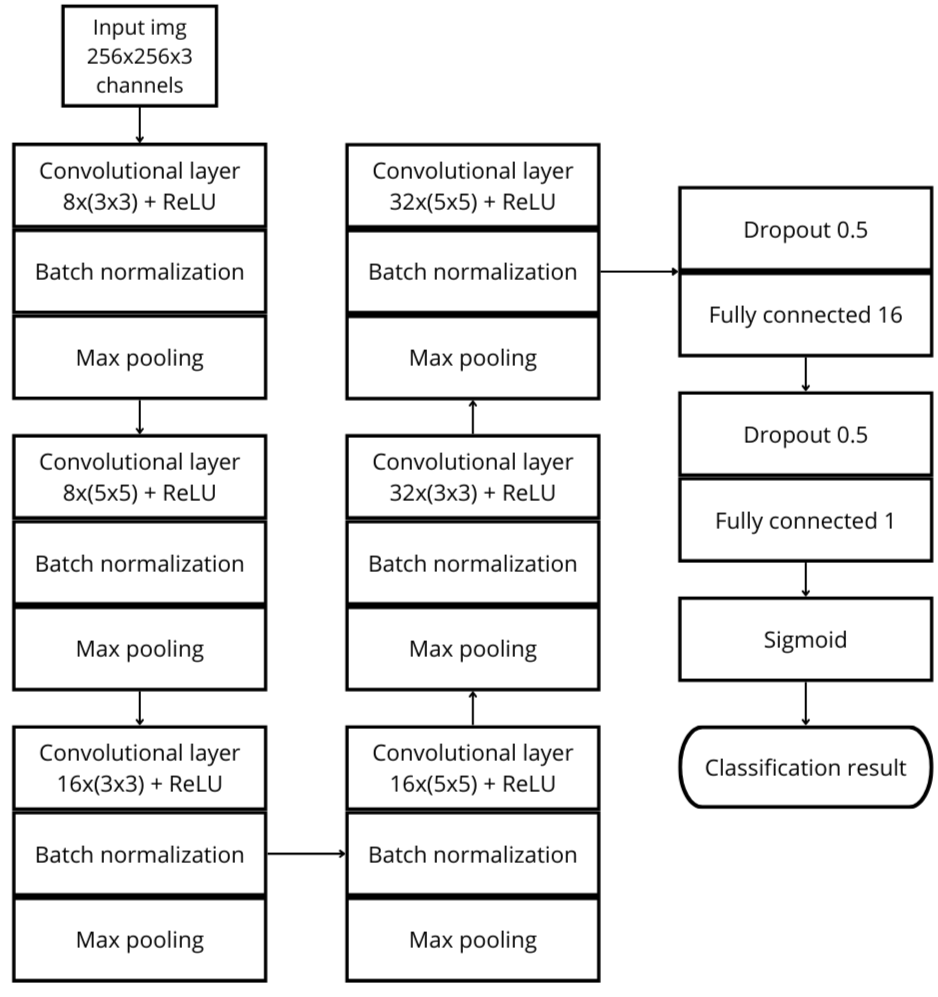}
    \captionof{figure}{6-layer convolutional structure of MesoNet+.}
    \label{fig:MesoNet+}
\end{figure}

Specifically, the new layers use 32 filters with a kernel size of  \(3 \times 3\), which serve to significantly enhance the model's feature extraction capabilities and boosting accuracy. Although this modification slightly increases computational costs, the resulting improvement in performance across diverse datasets justifies the trade-off. The refined architecture achieves greater precision in distinguishing between manipulated and authentic images.

The architecture follows a structured design, beginning with convolutional layers, each paired with batch normalization and max pooling. The convolutional layers extract spatial features from input images, identifying elements such as edges, artifacts, and textures. Batch normalization stabilizes and accelerates training by normalizing batch activations, while max pooling reduces feature dimensionality, improving robustness to minor variations and translations.

After the feature extraction phase, the activations are flattened and processed through dense layers. To mitigate overfitting, techniques such as Dropout are employed, randomly deactivating a portion of neural units during training. The fully connected layers, utilizing ReLU activations, model nonlinear relationships within the extracted features. Finally, the Softmax function in the output layer generates a probability distribution over the classes, enabling effective multiclass classification.

\subsection{Multi-label classification system. MesoMultiNet}

The primary focus of this work is developing a multi-class classification system capable of managing multiple categories, thereby extending its applicability across various image classification scenarios.

The MesoMultiNet architecture is a convolutional neural network (CNN) composed of several essential layers, including convolutional layers, batch normalization, max pooling, and dense layers. To facilitate multi-class classification, the network's output is configured to generate probabilities across multiple categories using a Softmax activation function. This architecture, inspired by the original MesoNet, is extended with four convolutional layers designed for multi-class classification tasks.

The training process for the multi-class model involves several well-defined steps. First, image data are loaded and preprocessed. The preprocessing implicates scaling the images to a normalized range of [0, 1] using a data generator and organizing them into batches to optimize the training process. Next, the multiclass model is defined and compiled. The model is constructed using the aforementioned architecture and compiled with the Adam optimizer, employing a specified learning rate. Categorical cross-entropy is employed as the loss function, particularly suited for multi-class classification tasks.

Training begins by feeding the preprocessed data into the model over multiple epochs. During each epoch, the model iteratively adjusts its internal weights to minimize the defined loss function, improving its predictive performance.

A critical element of the system is the transfer learning process from a pre-trained binary model to a multi-class model. This involves several stages. First, the pre-trained binary model, Meso4Binary, is loaded to leverage its existing trained weights. These weights, corresponding to the convolutional and normalization layers of the binary model, are transferred to the multiclass model, providing a robust foundation of learned features. This step significantly enhances the multi-class model’s ability to adapt to the more complex classification task.

Subsequently, adjustments are made to the dense layer weights. The weights from the binary model’s dense layer are adapted to align with the multiclass model’s dense layer, redistributing the weights of the two binary classes (Deepfake and Real) across the three multiclass categories (Deepfake, Real, and FaceSwap). This adaptation equips the model to effectively differentiate between the three classes of interest in the multi-class classification context.

After training, the model is evaluated using standard classification metrics, including the classification report and confusion matrix. These metrics offer a comprehensive assessment of the model's performance, detailing accuracy, recall, and F1-score for each class. The confusion matrix provides a visual representation of correct and incorrect predictions, highlighting specific areas where the model may require further refinement.

\subsection{Multiple classification system with 6 convolutional layers.
MesoMultiNet+}

Following the results of MesoMultiNet, the objective is to obtain a model capable of generalizing and detecting different types of images with greater accuracy. In this new architecture, we will increase the number of convolutional layers from 4 to 6, increasing the number of filters and distributing the layer size to obtain a better ability to detect edges, details, and general characteristics of the different classes.

\section{Evaluation of image classification models. Results and comparative analysis}\label{sec:exp}

This section presents a detailed analysis of the progress and evolution of the convolutional network based on the MesoNet architecture, a classifier network for genuine images and digital image manipulations. It has been segmented at various points to provide a comprehensive overview of the development and improvements incorporated in each successive version. This approach allows a full understanding of how the network has evolved over time and how the challenges and limitations encountered in previous versions have been addressed.

\subsection{Dataset}

In this research, the dataset plays a fundamental role in training and evaluating the proposed models in the context of detecting and classifying images generated by DeepFake techniques. To ensure the validity and robustness of the experiments, we selected and configured state-of-the-art databases, complemented with specific modifications tailored to meet the particular needs of our system. Below, we detail the selection, organization, and use of the various datasets employed in different phases of the study.

Dataset created by MesoNet principal investigators, images generated by DeepFake methodologies. For the first response, the following data samples are studied\cite{mesonet}. In a first phase, the model was adapted to work with a limited number of predictions, thus avoiding that the memory buffer of our computer would fill up and causing the interruption of the evaluation. For this purpose, in this first phase, we limited ourselves to a total of 19509 images divided into training and evaluation sets in Table \ref{tab:Tabla distribucion MesoNet DB}.

\begin{table}[h]
    \centering
    \begin{tabular}{|l|c|c|}
        \hline
        Set & Set deepfake images & Set real images \\
        \hline
        Train & 5111 & 7250 \\
        Validation & 2889 & 4259 \\
        \hline
    \end{tabular}
    \caption{Distribution of the training and evaluation dataset}
    \label{tab:Tabla distribucion MesoNet DB}
\end{table}

After the first evaluation with the original MesoNet data, a dataset of images not known to the model is run to verify its performance, this can be seen in Table \ref{tab: Celeb-DF frames&Videos}, using the widely known basis in the state-of-the-art, Celeb-DF\cite{li2020celeb}.

\begin{table}[h]
    \centering
    \begin{tabular}{|l|c|c|}
        \hline
        Celeb-DF & DeepFake & Real \\
        \hline
        Videos & 795 & 158 \\
        \hline
        Extracted images & 5642 & 1107 \\
        \hline
    \end{tabular}
    \caption{Distribution of Celeb-DF evaluation dataset}
    \label{tab: Celeb-DF frames&Videos}
\end{table}

Once the binary model was evaluated, with the known and unknown image datasets from the training process, we performed a final balanced dataset between the original\cite{mesonet} and Celeb-DF\cite{li2020celeb} databases to verify the performance of the system, can be seen in Table \ref{tab:Tabla distribucion MesoNet Balanceada}.

\begin{table}[h]
    \centering
    \begin{tabular}{|l|c|c|c|}
        \hline
        Dataset & Database original & Celeb-DF & Total \\
        \hline
        DeepFake images & 2845 & 2042 & 4887 \\
        Bonafide images & 4259 & 1101 & 5360 \\
        \hline
    \end{tabular}
    \caption{Distribution of the balanced dataset to evaluate the MesoNet model}
    \label{tab:Tabla distribucion MesoNet Balanceada}
\end{table}

For the experiments of this binary 6-layer convolutional model, we use the balanced dataset we created for the last evaluation of the MesoNet model, can be seen in Table \ref{tab:Tabla distribucion MesoNet Balanceada}.

In the multi-classification system, a public database containing images of FaceSwap DFDC attacks is used for the evaluation of the system, first of all a public database containing FaceSwap DFDC attack images \cite{dolhansky2020deepfake}. Both genuine images and DeepFake images belong to the Celeb-DF database\cite{li2020celeb}.

The database used to train and evaluate the multi-classifier system has been balanced, ensuring an equal distribution of samples between the different classes, this can be seen in Table \ref{tab:Tabla distribucion MesoMultiNet DB}. This ensures that the model is trained and evaluated fairly and effectively in all categories.

\begin{table}[h]
    \centering
    \begin{tabular}{|l|c|c|c|c|}
        \hline
        Dataset & Bonafide & DeepFake & FaceSwap & Total \\
        \hline
        Imágenes & 1 483 & 1 538 & 1 450 & 4 471\\
        \hline
    \end{tabular}
    \caption{Distribution of the dataset for the evaluation of MesoMultiNet and the first evaluation of MesoMultiNet+}
    \label{tab:Tabla distribucion MesoMultiNet DB}
\end{table}

 For the multiple classification model with 6 convolutional layers MesoMultiNet+, the same data set is used as in the 4 convolutional layers model MesoMultiNet+. The balanced data set defined above is used for this experiment, can be seen in Table \ref{tab:Tabla distribucion MesoMultiNet DB}.

When comparing FaceSwap images with DeepFake images, we found that these two classes are complex to differentiate, so we made a change to the dataset.  With respect to the images of genuine users, the samples used in the previous set are maintained, as are the attacks classified as FaceSwap \cite{li2020celeb}. The DeepFake images from the previous set are replaced by images generated with Stable Diffusion from the DifussionDB dataset \cite{wang2022diffusiondb}. The distribution chart of the dataset can be seen in Table \ref{tab:Tabla distribucion MesoMultiNet+ con SD}. 

\begin{table}[h]
    \centering
    \begin{tabular}{|l|c|c|c|c|}
        \hline
        Dataset & Bonafide & DeepFake & FaceSwap & Total \\
        \hline
        Images & 1 489 & 1 550 & 1 441 & 4 480\\
        \hline
    \end{tabular}
    \caption{Distribution of the MesoMultiNet+ deepfake dataset generated with Stable diffusion included in the set.}
    \label{tab:Tabla distribucion MesoMultiNet+ con SD}
\end{table}

\subsection{Preliminary Outcomes and Model Adaptations of MesoNet}

The choice of using this convolutional neural network was due to its reduced computational cost, being able to run on machines with limited computational capacity. 

For the representation of the results obtained after a first evaluation, the ROC curve is used. It represents the relationship between the true positive rate TPR and the false negative rate FPR at different decision thresholds. It is a useful metric for evaluating the performance of a binary classifier on a balanced data set. The AUC-ROC is interpreted as the probability that a model correctly classifies a positive instance, in this case a deepfake, and a negative instance, a genuine, selected at random. In the ROC curve, the x-axis represents the false positive rate FPR, which is calculated as the number of false positives divided by the sum of true negatives and false positives, and the y-axis represents the true positive rate TPR, which is calculated as the number of true positives divided by the sum of true positives and false negatives.

Ideally, the ROC curve is intended to be as close as possible to the upper left-hand corner of the graph, indicating a high rate of true positives and a low rate of false positives, regardless of the decision threshold, an ROC curve that is closer to the upper left-hand corner represents better model performance. The area under the ROC curve, AUC-ROC, is a quantitative measure of model performance, ranging from 0 to 1, where a value of 1 indicates a perfect model and a value of 0.5 indicates a performance similar to chance. In general, the higher the value of AUC-ROC, the better the performance of the model in classifying positive and negative classes.

\begin{figure}[htpb]
    \centering
    \includegraphics[width=0.5\textwidth]{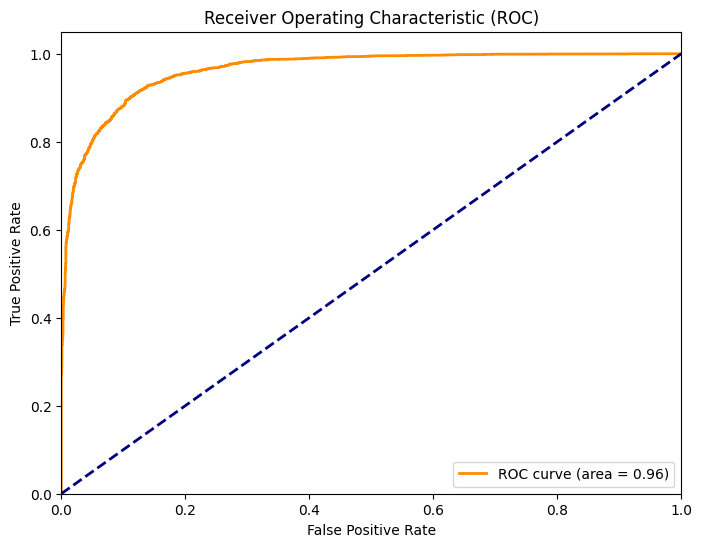}
    \captionof{figure}{True positive rate and false negative rate in the evaluation of the balanced MesoNET model.}
    \label{fig:ROCv1}
\end{figure}

In this first evaluation we can see how by evaluating the samples in the database we obtain extraordinary results, an AUC-ROC value of 0.96 indicates that the model is very good at distinguishing between deepfakes and genuine images in the database used for the evaluation this can be seen in Table \ref{tab:Tabla distribucion MesoNet DB}. In other words, the model is able to correctly classify most of the instances in the database.

The graph shown above that the performance of this binary classification model works with high levels of security and discrimination Fig. \ref{fig:ROCv1}. A second evaluation is carried out with a dataset used in this field of research Celeb-DF. This database is composed of videos with audiovisual content of genuine individuals and digital manipulations, categorized as DeepFake, without specifying the type of manipulation\cite{li2020celeb}. For the processing of these data, frames have been extracted from the videos in order to keep a reduced number of them, extracting one frame per second and obtaining the number of images detailed, can be seen in Table \ref{tab: Celeb-DF frames&Videos}

The convolutional model performs an evaluation of square images with dimensions of 256 pixels width and height. Therefore, an image alignment and matching algorithm has been employed, which can be seen in. After providing an unknown data set and performing the necessary preprocessing, an imprecise performance of the classification system is evident. The original accuracy, which stood at 0.96, decreases drastically to 0.56. An AUC-ROC value of 0.56 indicates that the model fails to effectively differentiate between deepfakes and real images. This scenario suggests the presence of overfitting or lack of generalization in the model, which implies its inability to generalize correctly to new instances that were not part of the training set.

\begin{figure}[htpb]
    \centering
    \includegraphics[width=0.5\textwidth]{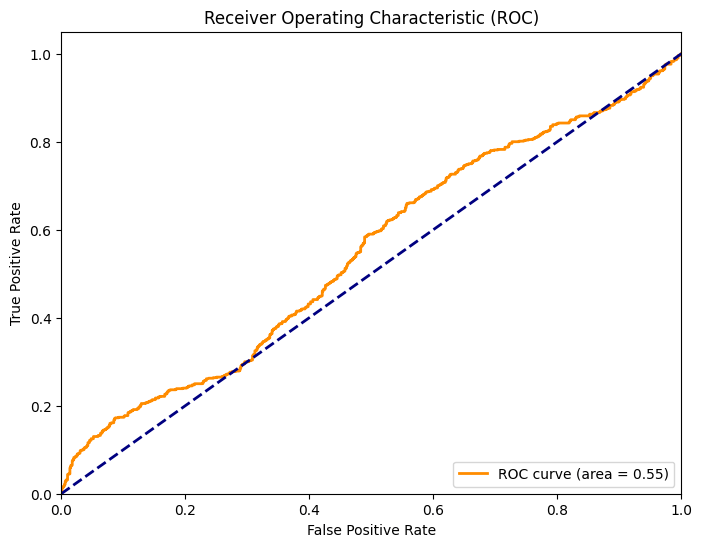}
    \captionof{figure}{True positive rate and false negative rate in the evaluation of the MesoNET model using an unknown dataset.}
    \label{fig:ROC 0.55}
\end{figure}

\begin{figure}[htpb]
    \centering
    \includegraphics[width=0.5\textwidth]{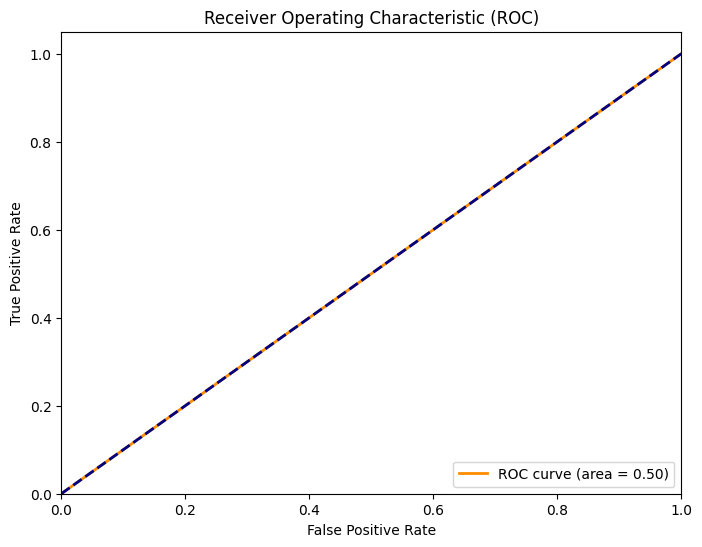}
    \captionof{figure}{True positive rate and false negative rate in the evaluation of the MesoNET model using a balanced dataset.}
    \label{fig:ROC 0.5}
\end{figure}

In order to solve this problem in the binary classification, two combined solutions were used to increase the level of accuracy, resulting in MesoNet+, with a balanced dataset and an increased convolution layer, as developed in the section, can be seen in Table \ref{tab:Tabla distribucion MesoNet Balanceada}. Prior to the development of our new MesoNet+ model, the new balanced dataset was evaluated, seeing that it worsened the previous results of unknown images Fig. \ref{fig:ROC 0.5}.

A ROC value of 0.50 indicates that the model does not have the ability to discriminate between the real and DeepFake classes. This may be due to the fact that the deconvolution layers do not have the ability to discern sufficient image characteristics in the evaluation process. This is solved in the MesoNet+ binary model.

\subsection{Evaluation and analysis of the results of the MesoNet+ binary classification model.}

The binary classification model we have developed allows us to obtain better results with unknown images. Prior to the new evaluations with the balanced set, the improvement process has been detailed in the previous section.

\begin{figure}[htpb]
    \centering
    \includegraphics[width=0.5\textwidth]{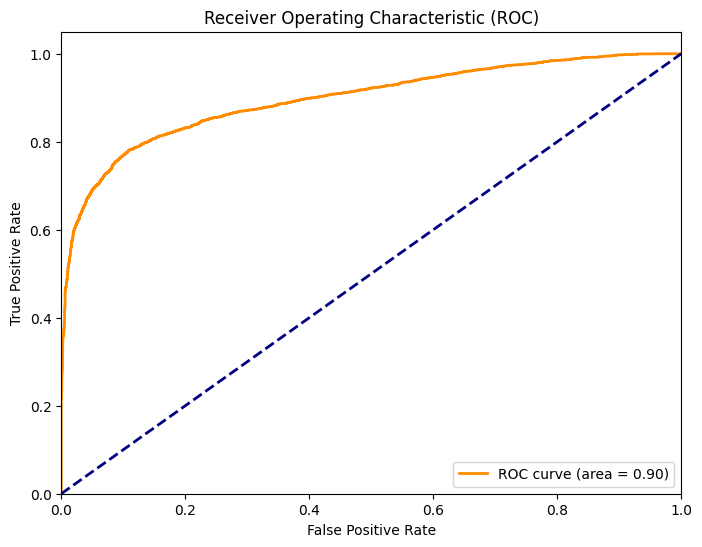}
    \captionof{figure}{True positive rate and false negative rate in the evaluation of the MesoNET+ model.}
    \label{fig:ROC MesoNet+}
\end{figure}

After implementing the changes in the architecture and in the dataset to be evaluated, the level of accuracy in the binary classification reaches 90\% accuracy for a balanced dataset Fig \ref{fig:ROC MesoNet+}.

The significant improvement in predicting DeepFake and real images can be seen due to the incorporation of the new deconvolution layers. This increase in accuracy also leads to an increase in computational expenditure. 

\subsection{Evaluation and performance analysis of the MesoMultiNet multi-class classification model}

The multi-class classification system was the objective proposed at the beginning of this work, having the ability to discriminate between different attacks on image manipulations and genuine users.

The analysis begins with the discrimination between genuine users and DeepFake and FaceSwap attacks. A balanced database is employed for training and evaluation. The model was adapted to develop a multi-classification system. Three distinct groups were identified.

\begin{figure}[htpb]
    \centering
    \includegraphics[width=0.5\textwidth]{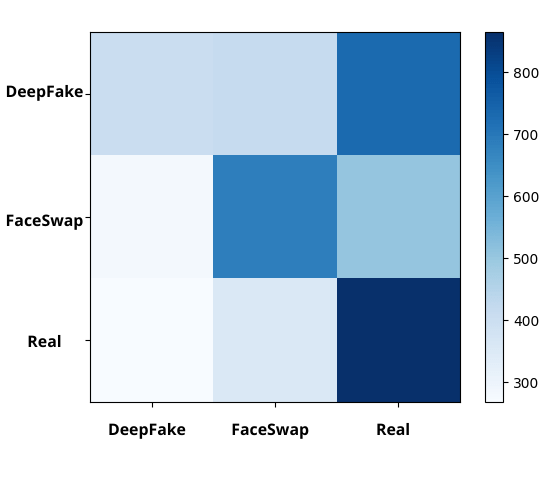}
    \captionof{figure}{MesoMultiNet architecture confusion matrix.}
    \label{fig:MatrizConfusiónErronea}
\end{figure}

After training the multiclassification model and performing the evaluation we can see how the classification is not performed correctly Fig. \ref{fig:MatrizConfusiónErronea}.

The performance metrics for each class in a multi-class classification model, as well as the overall averages of the metrics.

\begin{table}[h!]
\centering
\begin{tabular}{|l|c|c|c|c|}
\hline
\textbf{Class} & \textbf{Precision} & \textbf{Recall} & \textbf{F1 Score} & \textbf{Support} \\
\hline
DeepFake & 0.43 & 0.26 & 0.32 & 1550 \\
FaceSwap & 0.47 & 0.47 & 0.47 & 1465 \\
Real & 0.41 & 0.58 & 0.48 & 1488 \\
\hline
\textbf{Accuracy} & \multicolumn{4}{|c|}{0.43} \\
\hline
\textbf{Macro average} & 0.44 & 0.44 & 0.42 & 4503 \\
\hline
\end{tabular}
\caption{Performance metrics of the multi-class classification model}
\label{table:metrics}
\end{table}

For the DeepFake class (labelled 0), the accuracy, the 43\% of predictions are correct, while the recall, the 26\% of actual DeepFakes were identified. The F1 score for this class is 0.32, with a support of 1550 instances in the dataset.

For the FaceSwap class (labelled 1), the accuracy and recall are 47\% of predictions and actual FaceSwaps were correctly identified. The F1 score is 0.47, with a support of 1465 instances.

For the Real class (labelled 2), the accuracy is 0.41, and the recall is 0.58, indicating that 41\% of predictions and 58\% of actual Real images were correct. The F1 score is 0.48, with a support of 1488 instances.

The model's overall accuracy is 43\% predictions are correct. The unweighted averages for accuracy, recall, and F1 score are 0.44, 0.44, and 0.42, respectively. When weighted by class support, the averages are 0.44 for accuracy, 0.43 for recall, and 0.42 for F1 score.

These metrics highlight the model's strengths and limitations. FaceSwap demonstrates a more balanced performance in accuracy and recall compared to DeepFake and Real, suggesting greater effectiveness in identifying FaceSwap instances. However, the model's overall performance is hindered by two main factors: a limited dataset size of approximately 1500 images per class, which impacts generalization, and the similarity between FaceSwap and DeepFake images. Additionally, the use of a four-layer convolutional architecture, based on the original MesoNet, further restricts the model's ability to distinguish fine-grained features.

To address these challenges, the MesoMultiNet+ architecture was developed, increasing the number of convolutional layers from 4 to 6. This enhancement, inspired by the transition from MesoNet to MesoNet+, aims to improve the model's ability to detect and generalize features, leading to better classification performance.

\subsection{Evaluation and performance analysis of the MesoMultiNet+ multiclass classification model}

Following the modification of the convolutional layers, which increased the computational cost during training and evaluation, the added layers with a higher number of filters and the use of variable filter sizes—smaller filters for detail and edge detection and larger filters for more general feature extraction—produced significantly improved results. The confusion matrix obtained after the architectural changes is presented.

\begin{figure}[htpb]
    \centering
    \includegraphics[width=0.5\textwidth]{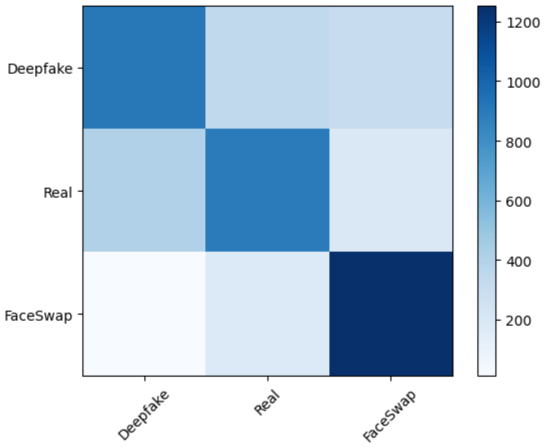}
    \captionof{figure}{MesoMultiNet+ architecture confusion matrix.}
    \label{fig:MatrizConfusiónSinteticosMMN+}
\end{figure}

The performance metrics for each class in a multi-class classification model, as well as the overall averages of the metrics, this can be seen in Table \ref{table:Tabla métricas MesoMultiNet+ sin sintéticos}.

\begin{table}[h!]
\centering
\begin{tabular}{|l|c|c|c|c|}
\hline
\textbf{Class} & \textbf{Precision} & \textbf{Recall} & \textbf{F1 Score} & \textbf{Support} \\
\hline
DeepFake & 0.68 & 0.59 & 0.63 & 1550 \\
FaceSwap & 0.72 & 0.87 & 0.78 & 1441 \\
Real & 0.63 & 0.60 & 0.62 & 1489 \\
\hline
\textbf{Accuracy} & \multicolumn{4}{|c|}{0.68} \\
\hline
\textbf{Macro average} & 0.68 & 0.68 & 0.68 & 4480 \\
\hline

\end{tabular}
\caption{MesoMultiNet+ multi-class classification model performance metrics}
\label{table:Tabla métricas MesoMultiNet+ sin sintéticos}
\end{table}

For the DeepFake class, the accuracy is 68\% of predictions labeled as DeepFake are correct. The recall 59\% of actual DeepFakes were identified. The F1 score is 0.63, with a support of 1550 images.

For the FaceSwap class, the accuracy is 72\% of predictions correctly labeled as FaceSwap. The recall is 87\% of actual FaceSwaps were identified. The F1 score is 0.78, with a support of 1441 images.

For the Real class, the accuracy, with 63\% of predictions correctly labeled as Real. The recall is 60\% of actual Real images were identified. The F1 score is 0.62, with a support of 1489 images.

The model's overall accuracy is 68\% of all predictions are correct. The unweighted averages for accuracy, recall, and F1 score are all 0.68, calculated as simple averages across the classes. The weighted averages, accounting for class support, are also 0.68 for all three metrics.

These metrics highlight the model's improved performance, particularly for the FaceSwap class, which achieves the highest accuracy, recall, and F1 score. This indicates greater effectiveness in identifying FaceSwap instances compared to DeepFake and Real.

Comparing these results to the 4-layer architecture, the MesoMultiNet+ model with 6 convolutional layers demonstrates a clear improvement. This is evident when analyzing the metrics in the MesoMultiNet+ results, can be seen in Table \ref{table:Tabla métricas MesoMultiNet+ sin sintéticos} versus those in the MesoMultiNet results, this can be seen in Table \ref{tab:Tabla distribucion MesoNet DB}.

\subsection{Training and evaluation of the MesoMultiNet+ model with the inclusion of AI-generated images}

In the latter proposed multi-classification architecture, results of 76\% accuracy are obtained.

\begin{figure}[htpb]
    \centering
    \includegraphics[width=0.5\textwidth]{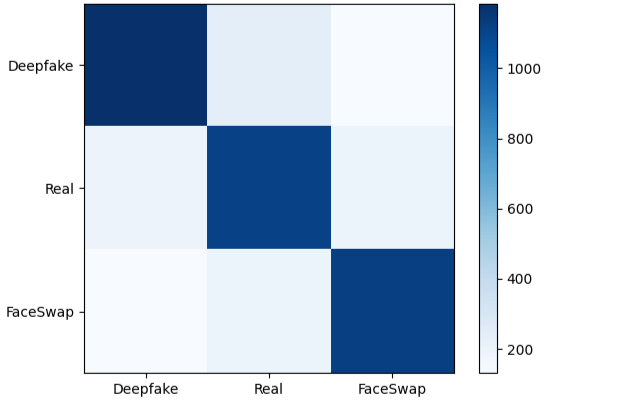}
    \captionof{figure}{MesoMultiNet+ architecture confusion matrix.}
    \label{fig:MatrizConfusiónMMN+}
\end{figure}

The performance metrics for each class in a multi-class classification model, as well as the overall averages of the metrics, can be seen in Table \ref{table:Tabla de resultados MMN+ con SD}.

\begin{table}[h!]
\centering
\begin{tabular}{|l|c|c|c|c|}
\hline
\textbf{Class} & \textbf{Precision} & \textbf{Recall} & \textbf{F1 Score} & \textbf{Support} \\
\hline
DeepFake & 0.79 & 0.76 & 0.78 & 1550 \\
FaceSwap & 0.78 & 0.78 & 0.78 & 1441 \\
Real & 0.73 & 0.75 & 0.74 & 1489 \\
\hline
\textbf{Accuracy} & \multicolumn{4}{|c|}{0.76} \\
\hline
\textbf{Macro average} & 0.76 & 0.76 & 0.76 & 4480 \\
\hline
\end{tabular}
\caption{Table of results MMN+ with Stable Diffusion.}
\label{table:Tabla de resultados MMN+ con SD}
\end{table}

For the DeepFake class, the accuracy, the 79\% of the predictions labelled as DeepFake are correct. The completeness, or recall, the that 76\% of all real DeepFakes were correctly identified. The F1 score, which is the harmonic mean between accuracy and completeness, is 0.78. The support for this class is 1550 images, which represents the total number of real DeepFake occurrences in the dataset.

For the FaceSwap class, the accuracy, the 78\% of the predictions labelled as FaceSwap are correct. The completeness is 0.78, which means that 78\% of all real FaceSwaps were correctly identified. The F1 score is 0.78. The support for this class is 1441 images.

For the Real class, the accuracy is 0.73, indicating that 73\% of the predictions labelled as Real are correct. The completeness is 0.75, which means that 75\% of all Real reals were correctly identified. The F1 score is 0.74. The support for this class is 1489 images.

The overall accuracy of the model is 0.76, meaning that 76\% of all predictions are correct regardless of class. The average of the metrics is 0.76 for accuracy, 0.76 for completeness and 0.76 for F1 score, calculated as the simple average of the metrics without weighting by the support of each class. The weighted average of the metrics is 0.76 for accuracy, 0.76 for completeness and 0.76 for F1 score, calculated weighted by the support of each class.

These metrics provide a complete picture of the model's performance in each class and overall, allowing areas where the model may need improvement to be identified. For example, the FaceSwap and DeepFake classes have a more balanced performance in terms of accuracy and completeness compared to Real, suggesting that the model is more effective at correctly identifying instances of FaceSwap and DeepFake.

\subsection{Comparison of multi-class classification models}

Table \ref{table:compact_vertical_comparison_metrics} presents a detailed comparison of the performance metrics for the MultiMesoNet, MultiMesoNet+, and MultiMesoNet+ with Stable Diffusion architectures.

The results highlight the progressive improvements achieved through architectural enhancements. MultiMesoNet+ exhibits a significant increase in accuracy 68\% compared to MultiMesoNet 43\%, attributed to the addition of layers that improve feature extraction. Furthermore, integrating Stable Diffusion images further enhances performance, achieving the highest accuracy 76\% and balanced metrics across all classes. This demonstrates the robustness of the enhanced architectures in distinguishing manipulated images with greater precision and consistency.

\begin{table}[h!]
\centering
\begin{tabularx}{\linewidth}{|l|X|X|X|}
\hline
\textbf{Metric} & \textbf{MultiMesoNet} & \textbf{MultiMesoNet+} & \textbf{MMN+ w/ SD} \\ \hline
\multicolumn{4}{|c|}{\textbf{DeepFake}} \\ \hline
Precision & 0.43 & 0.68 & 0.79 \\ \hline
Recall & 0.26 & 0.59 & 0.76 \\ \hline
F1 Score & 0.32 & 0.63 & 0.78 \\ \hline
\multicolumn{4}{|c|}{\textbf{FaceSwap}} \\ \hline
Precision & 0.47 & 0.72 & 0.78 \\ \hline
Recall & 0.47 & 0.87 & 0.78 \\ \hline
F1 Score & 0.47 & 0.78 & 0.78 \\ \hline
\multicolumn{4}{|c|}{\textbf{Real}} \\ \hline
Precision & 0.41 & 0.63 & 0.73 \\ \hline
Recall & 0.58 & 0.60 & 0.75 \\ \hline
F1 Score & 0.48 & 0.62 & 0.74 \\ \hline
\multicolumn{4}{|c|}{\textbf{Overall Metrics}} \\ \hline
Accuracy & 0.43 & 0.68 & 0.76 \\ \hline
Macro Avg. Precision & 0.44 & 0.68 & 0.76 \\ \hline
Macro Avg. Recall & 0.44 & 0.68 & 0.76 \\ \hline
Macro Avg. F1 Score & 0.42 & 0.68 & 0.76 \\ \hline
Support & 4503 & 4480 & 4480 \\ \hline
\end{tabularx}
\caption{Comparison of performance metrics for MultiMesoNet, MultiMesoNet+, and MultiMesoNet+ with Stable Diffusion.}
\label{table:compact_vertical_comparison_metrics}
\end{table}

\section{Conclusions}\label{sec:conclusion}

In this study, various convolutional neural network (CNN) architectures were designed and evaluated for the classification of digitally manipulated images. The research yielded promising results through the development and implementation of classification models optimized for low computational cost, enabling a discriminative model suitable for execution in resource-constrained systems.

The development of the MesoNet+ and MesoMultiNet+ architectures has demonstrated high efficiency in classifying genuine images versus manipulated images. Especially, the MesoMultiNet+ model stands out for its ability to improve performance by including AI-generated images in its training. This underlines the importance of using varied and representative data sets during model training to capture the diversity of possible manipulations.

Compared to other existing models, MesoMultiNet+ has a slightly better performance, as evidenced by metrics such as accuracy, completeness and F1 score. This model is particularly effective in identifying images manipulated by advanced techniques such as DeepFake and FaceSwap, indicating its potential for practical applications in real scenarios, in low-cost systems.

With respect to accuracy, the models have achieved high levels of discrimination in the detection of manipulated images. However, it is observed that the Real class has a slightly lower performance, suggesting the need for further improvements in the discrimination of genuine images.

Recall metrics reflect a good balance in the identification of different types of image manipulations. This balance is crucial for practical applications where both false positive and false negative detection can have significant implications.

In reference to data preprocessing, preprocessing techniques have played a crucial role in improving model performance. Adequate preprocessing of the images before their input into the neural network has improved the quality of the data and reduced noise, which implies a higher accuracy and robustness of the models.

The use of transfer learning allows the training process to be accelerated and results to be significantly improved by exploiting prior knowledge gained from pre-trained models on large data sets. This technique proves particularly useful for tackling complex problems with limited data.

\section*{Acknowledgments}

We would like to thanks Facephi for its support. This work has been funded by the Valencian regional
government CIAICO/2022/132 Consolidated group project
AI4Health. This work has also been supported by a Spanish national and a regional grant for PhD studies, FPU21/00414, and CIACIF/2022/175.

\bibliographystyle{IEEEtran}
\bibliography{biblio.bib}

\end{document}